\pdfoutput=1

\documentclass[11pt]{article}

\usepackage[final]{acl}
\usepackage{amsmath}

\usepackage{times}
\usepackage{latexsym}
 \usepackage{pifont}
 \usepackage{braket}

 \usepackage{CJKutf8} 

\usepackage[T1]{fontenc}

\usepackage[utf8]{inputenc}

\usepackage{microtype}

\usepackage{inconsolata}

\usepackage{graphicx}

\title{QCoder Benchmark: Bridging Language Generation and Quantum Hardware through Simulator-Based Feedback}

\author{%
Taku Mikuriya$^{1,2}$, 
Tatsuya Ishigaki$^{1}$, 
Masayuki Kawarada$^{1,3}$, 
{\bf Shunya Minami}$^{1}$, \\
{\bf Tadashi Kadowaki}$^{1,4}$, 
{\bf Yohichi Suzuki}$^{1}$, 
{\bf Soshun Naito}$^{5}$, 
{\bf Shunya Takata}$^{6}$, \\
{\bf Takumi Kato}$^{7}$,
{\bf Tamotsu Basseda}$^{8}$,
{\bf Reo Yamada}$^{9}$,
{\bf Hiroya Takamura}$^{1}$
\\
\\
$^{1}$National Institute of Advanced Industrial Science and Technology (AIST) \\
$^{2}$Yokohama National University\quad
$^{3}$CyberAgent, Inc.\quad
$^{4}$DENSO CORPORATION\\
$^{5}$The University of Tokyo~~
$^{6}$Keio University~~
$^{7}$NTT DATA GROUP Corporation~~\\
$^{8}$Miletos inc.~~
$^{9}$University of Tsukuba
}

\begin{document}
\maketitle
\begin{abstract}
Large language models (LLMs) have increasingly been applied to automatic programming code generation.
This task can be viewed as a language generation task that bridges natural language, human knowledge, and programming logic.
However, it remains underexplored in domains that require interaction with hardware devices, such as quantum programming, where human coders write Python code that is executed on a quantum computer.
To address this gap, we introduce QCoder Benchmark, an evaluation framework that assesses LLMs on quantum programming with feedback from simulated hardware devices.
Our benchmark offers two key features.
First, it supports evaluation using a quantum simulator environment beyond conventional Python execution, allowing feedback of domain-specific metrics such as circuit depth, execution time, and error classification, which can be used to guide better generation.
Second, it incorporates human-written code submissions collected from real programming contests, enabling both quantitative comparisons and qualitative analyses of LLM outputs against human-written codes.
Our experiments reveal that even advanced models like GPT-4o achieve only around 18.97\% accuracy, highlighting the difficulty of the benchmark. In contrast, reasoning-based models such as o3 reach up to 78\% accuracy, outperforming averaged success rates of human-written codes (39.98\%).
We release the QCoder Benchmark dataset along with a public evaluation API to support further research.\footnote{https://qcoder-bench.github.io/}
\end{abstract}

\section{Introduction}
\begin{CJK}{UTF8}{min}

Programming code generation has emerged as an important and practical problem in language generation studies~\cite{chen2021codex}.
This task requires models to generate correct and executable code by bridging natural language, human expertise, and formal programming logic.
Recent advances in large language models (LLMs) have led to impressive performance on classical programming benchmarks~\cite{wang2025openhands,openai2024gpt4technicalreport}.
However, these benchmarks are primarily evaluated in software-only environments, where failures are typically limited to runtime errors or syntax violations detected by a software development environment such as Python interpreters.
In contrast, little is known about how LLMs perform in domains such as quantum programming~\cite{vishwakarma2024qiskithumaneval}, where generated code must not only be syntactically correct, but also conform to strict, domain-specific constraints imposed by real or simulated quantum hardware.

\begin{figure} [t]
    \centering
    \vspace{2ex}
    \includegraphics[width=1.0\linewidth]{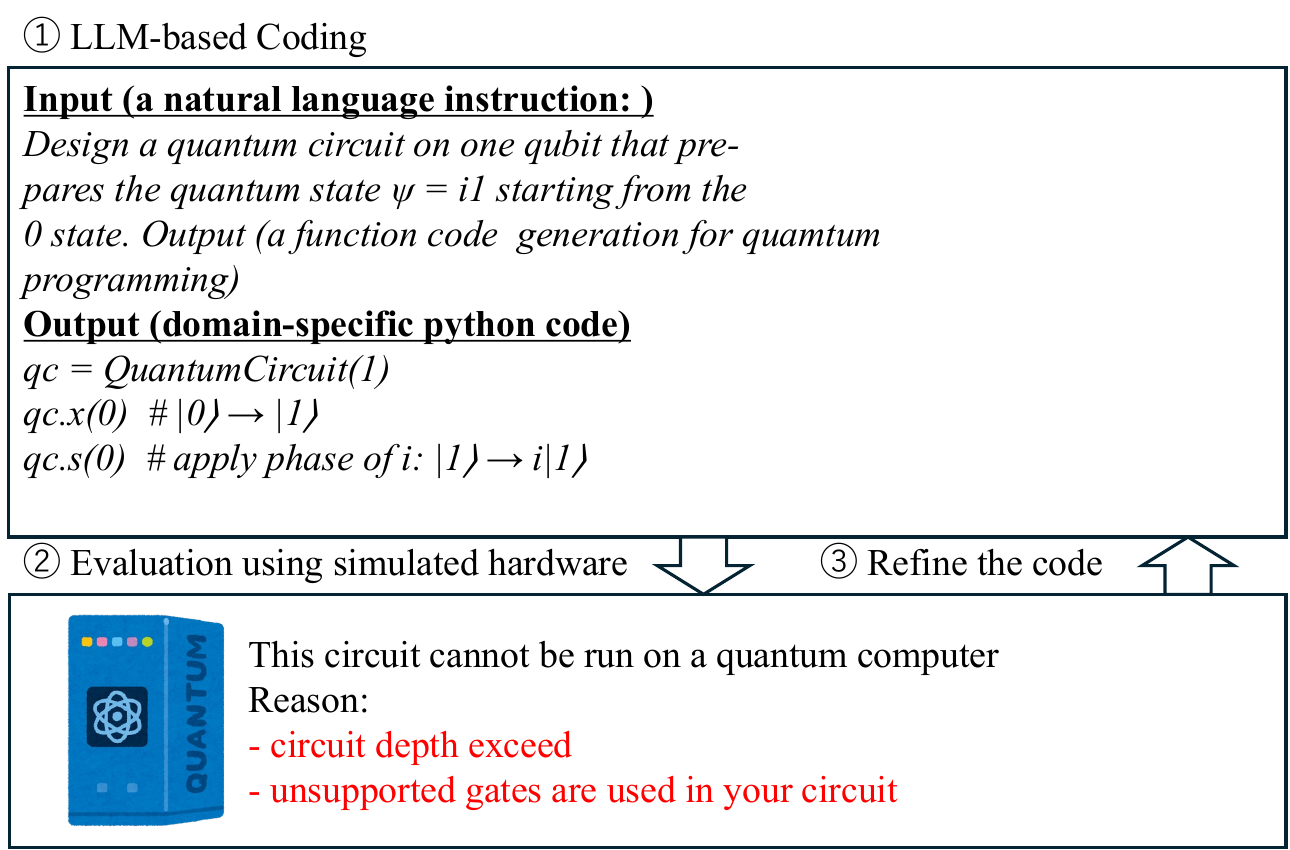}
    \caption{Quantum code generation involves generating a python code that constructs a quantum circuit executable on a quantum computer. Due to strict constraints of actual hardware, feedback from the hardware is necessary to generate executable codes on a quantum computer.}
    \label{fig:enter-label}
\end{figure}

Quantum programming serves as a representative example of hardware constraint-driven code generation tasks.
As shown in Figure~\ref{fig:enter-label}, given the instruction as a natural language, this code generation task involves generating a python code to produce a quantum circuit that can be run on a separate hardware, i.e., either a real quantum computer or simulator.
Unlike classical programs, correctness and executability depend not only on the absence of runtime errors in Python, but also on whether the resulting circuits comply with constraints imposed by quantum hardware.  
These hardware-level constraints include, for example, limitations on circuit depth (i.e., the number of sequential gate operations) and the availability of only certain types of quantum gates in a quantum computer.
As a result, quantum code must be syntactically valid in Python and must generate quantum circuits that conform to the logical requirements of quantum computation.


To enhance studies on this constrained generation task, we introduce QCoder Benchmark, a dataset and evaluation framework specifically designed for quantum code generation.  
Our benchmark contains 1) pairs of a programming contest problem and human-written solutions and 2) an evaluation tool to provide hardware-specific feedback. 
Unlike prior benchmarks that rely on generic Python execution, our evaluation tool uses a quantum simulator that returns quantum-specific feedback about domain-specific constrains, e.g., circuit depth and inappropriate uses of unsupported quantum gates.
This evaluation tool allows feedback-driven iterative language generation: a generation paradigm in which models incorporate feedback from a hardware to refine their generated codes~\cite{madaan2023selfrefine}.

This paper uses our benchmark to investigate whether LLMs can improve their quantum code generation performance by incorporating domain-specific feedback.  
Our experiments show that even advanced LLMs like GPT-4o achieve only around 18.97\% accuracy, while the best performing LLM o3 reaches 65.52\% and it outperforms averaged success rate of human-written codes submitted to programming contests (39.98\%).
We also find that incorporating feedback into prompt to refine codes can significantly improve generation performance, emphasizing the importance of feedback from a simulated hardware.

We release QCoder Benchmark as public resources to support further research on code generation under complex constraints.
This paper makes the following contributions: 1) we implement iterative code generators that use feedback from a simulated hardware-based evaluation tool, 2) we empirically demonstrate that such feedback effectively enhances LLMs' performances, and 3) our benchmark data and evaluation API will be made public.

\section{Related Work}

Various coding benchmarks have been proposed for general-purpose code generation tasks, such as HumanEval~\cite{chen2021codex}, Mostly Basic Python Problems (MBPP)~\cite{austin2021program}, and the APPS dataset~\cite{hendrycksapps2021}.  
These benchmarks primarily focus on solving basic algorithmic problems written in Python and are evaluated using predefined input-output test cases.

Our benchmark differs from these benchmarks in two key aspects.  
First, it targets a domain-specific coding task—quantum programming—which involves generating circuits that must conform to real-world hardware constraints.  
Second, rather than relying solely on static test cases, our benchmark evaluates generated code using a simulated quantum computer, offering a new paradigm for evaluating executable and hardware-aware code.

Domain-specific coding benchmarks have been introduced for various domains, including data science (DS-1000~\cite{Lai2022DS1000}), secure coding (LLMSecEval~\cite{llmseceval2023}), database query generation (Spider~\cite{yu-etal-2018-spider}), and bioinformatics (BioCoder~\cite{10.1093/bioinformatics/btae230}).

In the quantum programming domain, Qiskit HumanEval~\cite{vishwakarma2024qiskithumaneval} shares similar motivations with ours.  
However, our benchmark differs in two important ways: (1) it provides a hardware simulator-based evaluation framework for assessing generated quantum circuits, and (2) each programming task is accompanied by multiple human-written implementations, enabling comparative analysis between human and LLM-generated code.

Quantum programming is a form of code generation aimed at controlling hardware, but only a few attempts exist in this direction e.g., a study that generates code to control robots~\cite{LUO2024100488}.
Refinement of generated code has also been shown to be effective in various setups~\cite{refine_code,madaan2023selfrefine,bi-etal-2024-iterative,liu2023rltf}.
Our study extends this line of work by incorporating hardware-aware feedback.

\section{QCoder Benchmark}
\label{sec:qcoder_benchmark}

Our benchmark consists of pairs of programming contest problems and human-written solutions, together with an evaluation tool that provides quantum hardware-aware feedback~\footnote{This evaluation tool will be released as a Web API for easier usage for future researches.}.
Our benchmark differs from existing general-purpose or quantum benchmarks~\cite{vishwakarma2024qiskithumaneval} in two aspects: (1) it enables fine-grained evaluation from domain-specific perspectives (e.g., circuit depth), not just functional correctness, and (2) it includes human-written solutions collected from programming contest submissions.
These submissions often contain errors and variations useful for studying the differences between LLM-generated and human-written code.

\subsection{Dataset}

\subsubsection*{Formulation of Quantum Code Generation}

We define the task of quantum code generation as generating a quantum circuit implementation in response to a natural language prompt.
The input prompt describes a quantum programming problem in natural language, including constraints on hardware such as supported quantum gates or maximum circuit depth.
The expected output is a quantum program written using the Qiskit library~\cite{qiskit2024} (or a compatible framework) that can be transpiled into a valid quantum circuit.
An example of input prompt is shown in Figure~\ref{fig:prompt-formulation}.
\begin{figure}[thb]
\centering
\begin{tabular}{|p{0.95\linewidth}|}
\hline
\textbf{Problem} \\
Design a quantum circuit on one qubit that prepares the quantum state $\ket{\psi} = i\ket{1}$ starting from the $\ket{0}$ state. \\[1mm]

\textbf{Constraints} \\
States with different global phases will be considered incorrect. \\
Use the following code format: \\

\begin{verbatim}
from qiskit import QuantumCircuit

def solve() -> QuantumCircuit:
    qc = QuantumCircuit(1)
    # Write your code here:

    return qc
\end{verbatim}

The LLM is expected to generate only the body of the \texttt{solve()} function. \\
\hline
\end{tabular}
\caption{Example prompt for quantum code generation.}
\label{fig:prompt-formulation}
\end{figure}

\subsubsection*{Data Collection}

QCoder Benchmark is constructed by collecting quantum programming problems from QCoder, a publicly available platform for quantum programming education website\footnote{\url{https://www.qcoder.jp/en}}.
We have obtained permission from the QCoder's developers to redistribute the dataset.

Each problem includes a reference solution and the corresponding target quantum state vector.
Each problem is also paired with human-written solutions submitted by participants of real-world quantum programming contests hosted on QCoder.
Unlike many generation tasks where model outputs are compared to references using metrics such as BLEU~\cite{papineni-etal-2002-bleu}, code generation typically does not use references for evaluation, instead, we execute the generated code and verify functional correctness.

For each of the 58 problems, we collected approximately 30 human-submitted codes on average, resulting in a total of 1,740 problem–solution pairs. All solutions are written using the Qiskit library.
Although these 30 codes represent the final submitted versions, each was typically created through multiple rounds of editing and refinement by a human coder. The dataset also includes revision histories for each submission, with an average of 20 intermediate versions per code, capturing the iterative development process of human programmers.
This rich set of revisions reflects a diverse range of implementation strategies and coding styles, providing a challenging and realistic benchmark for LLM-based code generation.

\subsection{Simulator-based Evaluator}

Our benchmark has a simulator-based evaluation tool that assesses quantum codes.
This evaluator ensures that the submitted quantum circuits are not only functionally correct but also comply with hardware constraints.
Given a quantum program, the evaluator performs three evaluation steps:

\begin{enumerate}
    \item \textbf{Runtime check}: The program is executed using the Python interpreter to detect syntax or runtime errors. If an error occurs, the remaining evaluation steps are skipped.
    \item \textbf{Unsupported gates check}: If no runtime error is detected, the program is transpiled into a quantum circuit. The evaluator then checks whether any gates not supported by the hardware side are used. Such violations are critical, as they make the circuit non-executable on real quantum hardware~\footnote{In such cases, the code refinement can make the circuit be executable by decomposing unsupported gates into the supported gate set.}.
    \item \textbf{Circuit depth check}: The evaluation tool measures the circuit depth and compares it against the problem’s specified threshold, if any.
    \item \textbf{Output state fidelity check}: The circuit is executed on either a real quantum computer or a simulator, and the resulting quantum state is compared against the reference state to assess correctness.
    Note that our experiments use\footnote{\url{https://qiskit.github.io/qiskit-aer/}} simulator, but it can be also replaced by a real quantum computer for more precise feedback.
\end{enumerate}

This evaluation tool checks for constraint violations in order of severity from top to bottom.
From the software development environment, python's runtime errors are considered critical.
From the hardware side, the use of unsupported gates is treated as the most critical violation, as it renders the circuit incompatible with real quantum hardware. If a depth limit is specified in the prompt, depth violations are also flagged, as they may impact execution feasibility on NISQ (Noisy Intermediate-Scale Quantum) devices.

All evaluation steps are performed systematically on our web-based API, which will be made public. The API takes a generated quantum code as input and returns a textual report including runtime success, constraint violations, and correctness against the reference state represented in a predefined format, e.g, if an generated program is correct, the following report is produced: \texttt{\{ "runtime\_error": false, "gate\_violation": false, "depth\_violation": false, "state\_match": true \}}.
This feedback is converted into a natural language prompt used to refine generated code as explained in Section \ref{sec:methods}.

\subsection{Statistics and Comparisons with Other Datasets}

Table~\ref{tab:comparison} compares our QCoder Benchmark with existing code generation benchmarks. While prior datasets such as HumanEval~\cite{chen2021codex} and MBPP~\cite{austin2021program} focus on general programming tasks, QCoder is tailored for quantum programming as with an existing benchmark Qiskit HumanEval~\cite{vishwakarma2024qiskithumaneval}.
In contrast to Qiskit HumanEval~\cite{vishwakarma2024qiskithumaneval}, QCoder provides human-written solutions and hardware-aware evaluation tool.

\begin{table*}[t]
\centering
\small
\begin{tabular}{lccccp{4cm}}  
\hline
\textbf{Dataset} & \textbf{Domain} & \textbf{Submissions} & \textbf{Test Case} & \textbf{Hardware} & \textbf{Dataset Size} \\
\hline
HumanEval & General & N/A & Yes & N/A & 3,200 problem-answer pairs \\
MBPP & General & N/A & Yes & N/A & 1,000 problem-answer pairs \\
Qiskit HumanEval & Quantum Programming & N/A & Yes & Partial & 151 problem-answer pairs \\
\textbf{QCoder (Ours)} & Quantum Programming & \textbf{Yes} & \textbf{Yes} & \textbf{Yes} & 58 problems \newline $\times$ 30 human coders \newline $\times$ 20 submissions (on avg) \\
\hline
\end{tabular}
\caption{Comparison of QCoder Benchmark with existing code generation benchmarks. 
"Submissions" indicates whether the dataset includes multiple human-written solutions. 
"Test Case" refers to the use of predefined functional tests, 
and "Hardware" denotes whether the evaluation considers hardware-level constraints.}
\label{tab:comparison}
\end{table*}



\section{Methods}
\label{sec:methods}

This paper compares prompt-based code generators rather than finetuning-based models because developing prompt-based techniques is particularly important for domains like quantum programming, where users are often not experts in natural language processing.
The following subsections describe our prompt and refinement process using hardware-aware feedback expressed in natural language.

\subsection{Baseline Prompting Strategy}

For all LLMs used in our experiments, we use a consistent prompting strategy to ensure fair comparison. Each prompt includes:
\begin{itemize}
    \item A natural language problem description
    \item Explicit constraints (e.g., unsupported gates and depth constraints)
    \item A Python code template with a placeholder function \texttt{solve()} that uses the Qiskit library
\end{itemize}

Models are explicitly instructed to generate only the function body, with no additional imports or code outside the template.
A sample prompt is shown in Figure~\ref{fig:prompt-example}.

\begin{figure}[h!]
\centering
\begin{tabular}{|p{0.95\linewidth}|}
\hline
\textbf{Problem:} Design a quantum circuit on one qubit that prepares the quantum state $\ket{\psi} = i\ket{1}$ starting from $\ket{0}$. \\
\textbf{Constraints:} States with different global phases will be considered incorrect. \\
\textbf{Code template:} \\
\begin{verbatim}
from qiskit import QuantumCircuit
def solve() -> QuantumCircuit:
    qc = QuantumCircuit(1)
    # Write your code here
    return qc
\end{verbatim} \\
\hline
\end{tabular}
\caption{Example of the baseline prompt.}
\label{fig:prompt-example}
\end{figure}

We use the default tokenizer and decoding strategy of each model. No maximum token length is specified; decoding is stopped upon encountering a stop token or natural termination. The generated function body is inserted into the provided template and directly passed to the evaluation tool for assessment.

\subsection{Feedback-aware Code Refinement}

\begin{figure}[ht]
\centering
\begin{tabular}{|p{0.95\linewidth}|}
\hline
Your answer was \\
\begin{verbatim}
'''python
{LLMs' submitted code}
'''
\end{verbatim}
Branching according to labels: \\
\textbf{WA:} This is wrong. Try again. \\
\textbf{DLE:} The circuit depth exceeded the given constraint. Please revise your implementation to improve efficiency. Try again. \\
\textbf{UME:} Unauthorized modules has been used. Try again. \\
\textbf{UGE:} An unauthorized quantum gate has been used. Try again. \\
\textbf{RE:} The occurring error is: \{error text\}. Try again. \\
\hline
\end{tabular}
\caption{The prompt used for iterative refinement.}
\label{fig:prompt-refinement}
\end{figure}

In addition to the baseline prompting, we also evaluate an iterative refinement approach that utilizes feedback from the simulator-based evaluator described in Section~\ref{sec:qcoder_benchmark}.
This method aims to improve code correctness by performing multiple rounds of generation and correction.

As shown in the example prompt in Figure~\ref{fig:prompt-refinement}, at each iteration the model receives the baseline prompt along with structured feedback from the evaluator, such as Python error messages, circuit depth violations, or gate usage issues, in the predefined format explained in Section~\ref{sec:qcoder_benchmark}.
The model is instructed to revise its previous code while adhering to the original constraints.

This iterative process is repeated up to a fixed number of rounds (e.g., 3), or until the generated program passes all evaluation checks. This approach simulates a human-like refinement loop and allows us to assess whether LLMs can incorporate domain-specific feedback to improve their solutions over iterations.
This pipeline relies on the simulator-based evaluator introduced in Section~\ref{sec:qcoder_benchmark}, which serves both as a verifier and as a source of structured feedback during refinement.

\section{Experiments}

\subsection{Compared LLMs}
We evaluate three proprietary models: \texttt{gpt-3.5-turbo}, \texttt{gpt-4o-mini}, and \texttt{o3}.
We also compare two open-source models: \texttt{Qwen-1.5-14B-Chat}~\cite{qwen} and \texttt{DeepSeek-R1-Distill-Llama-70B}~\cite{deepseekai2025}.


These models were selected to cover both proprietary and open-source systems, as well as a range of model sizes and architectural designs, enabling broad comparisons of capabilities.
The proprietary models are accessed via the OpenAI API, while the open-source models are deployed locally with their default tokenizers and decoding configurations. All models are prompted using the same format and are evaluated under identical conditions using our simulator-based evaluator.

\subsection{Evaluation Metrics}

We use the following evaluation metrics to assess the generated programs:

\paragraph{Success rate.}
A generation is counted as successful if the produced code:
\begin{enumerate}
    \item runs without any Python execution errors,
    \item passes the simulator-based checks for unsupported gate usage and circuit depth,
    \item produces the correct quantum state vector as specified in the task.
\end{enumerate}

\paragraph{Fine-grained Failure Rates.}
To better understand the reasons behind failures, we compute the proportion of failed generations at each stage of the evaluation process. Specifically, we calculate the following stage-wise failure rates:

\begin{itemize}
\item \textbf{Runtime Error:} The proportion of generated programs that fail due to Python runtime errors.
\item \textbf{Gate Constraint Violation:} The proportion of programs that use quantum gates unsupported by the specified hardware.
\item \textbf{Depth Constraint Violation:} The proportion of programs whose circuit depth exceeds the specified limit.
\item \textbf{Wrong Output:} The proportion of programs that pass all checks but still produce an incorrect final quantum state vector.
\end{itemize}

These fine-grained metrics help isolate specific failure points and provide a more detailed characterization of model weaknesses.

\paragraph{Success Rates of Human Submitted Codes}
For each test problem, approximately 30 human-written code samples are available on average. We compute both the overall success rates and fine-grained failure rates on these human submitted solutions to investigate humans' coding skills.

\paragraph{Changes of Success Rates over iterations}
As the iterative feedback method generates new code at each round, we track the changes of success rate over iterations.



\section{Main Results}

In this section, we present the results of our experiments.
We begin by comparing overall success rates, followed by a fine-grained failure analysis to understand the types of errors commonly produced.
We then examine the effects of iterative refinement using simulator-based feedback and compare model performance against human-written code.

\subsection*{Which LLMs did work better?}

\begin{table}[t]
\centering
\small
\begin{tabular}{lc}
\hline
\textbf{Model} & \textbf{Success Rate (\%)} \\
\hline\hline
GPT-4o-mini     & 18.97 \\
GPT-3.5-turbo   & 10.34 \\
o3       & \textbf{65.52} \\ \hline
DeepSeek-R1-Distill-Llama-70B   & 29.31 \\
Qwen-1.5-14B-Chat   & 10.34 \\ \hline
Averaged Human & 39.98 \\
\hline
\end{tabular}
\caption{Success rate (\%) of baseline prompting without code refinements.}
\label{tab:zero-shot-results}
\end{table}

As shown in Table~\ref{tab:zero-shot-results}, o3, a reasoning-based model, achieved the highest success rate (\textbf{65.52}\%) among all compared LLMs.
It significantly outperforms other proprietary models: GPT-4o-mini (18.97\%), 
or GPT-3.5-turbo (10.34\%).
This result suggests the substantial advantage of reasoning-oriented models in quantum code generation.
As expected, open-sourced models, i.e., DeepSeek-R1-Distill-Llama-70B and Qwen-1.5-14B, obtain lower success rates than all proprietary models possibly due to smaller parameter sizes.

\subsection*{Fine-grained Failure Rates}

Next, we discuss failure rates in each evaluation step.


\begin{table*}[t]
\centering
\begin{tabular}{lcccc}
\hline
\textbf{Model} & \textbf{Success (\%)} & \textbf{Runtime Err. (\%)} & \textbf{Depth (\%)} & \textbf{Wrong Output (\%)} \\
\hline
GPT-4o-mini & 18.97 & 17.24 & 5.17 & 53.45 \\
GPT-3.5-Turbo & 10.34 & 36.21 & 12.06 & 31.03 \\
o3 & \textbf{65.52} & \textbf{0.00} & \textbf{1.72} & \textbf{18.97} \\ \hline
Averaged Human & 39.98 & 27.40 &  10.03 & 24.69 \\
\hline
\end{tabular}
\caption{Distribution of fine-grained failure rates identified by the hardware-aware evaluation tool when the number of iterations is set to one without refinement.}
\label{tab:failure-analysis}
\end{table*}

\noindent
\textbf{Comparing Among LLMs: }As shown in Table~\ref{tab:failure-analysis}, GPT-4o-mini and GPT-3.5-Turbo exhibit higher failure rates of runtime errors (17.24\% and 36.21\%, respectively), indicating frequent failures during the basic Python code execution stage.
Even when programs avoid runtime errors and meet the circuit depth constraints, a significant portion still fail to produce the correct quantum state vector upon simulation—accounting for 53.45\% in GPT-4o-mini and 31.03\% in GPT-3.5-Turbo.

In contrast, o3 demonstrates a remarkable reduction in runtime errors (0\%) and shows improved robustness across all error types. However, it still fails to produce the correct quantum output in 18.97\% of the cases, suggesting that despite being a reasoning-based model, there remains room for improvement in fully automating quantum programming via LLMs.

\noindent
\textbf{Comparing Human coders and LLMs: }Runtime Errors are common for both GPT-3.5-Turbo (36.21\%) and human coders (27.40\%), suggesting that generating syntactically valid codes remains a challenge for both.
For the depth violations, human coders (10.03\%) and GPT-3.5-Turbo (12.06\%) show similar violation rates, suggesting that understanding and complying with quantum-specific constraints such as circuit depth is equally challenging for both.
In contrast, o3 maintains a remarkably low violation rate (1.72\%), indicating its superior adaptation to quantum programming constraints.

Finally for the wrong outputs, the most prominent failure mode for GPT-4o-mini is generating incorrect output state vectors despite producing syntactically valid code.
While humans also exhibit a noticeable rate of wrong outputs (24.69\%), o3 performs better (18.97\%), suggesting its stronger task comprehension and ability to generate semantically accurate code.

\subsection*{How did iterative refinement improve generation?}

\begin{figure}
    \centering
    \includegraphics[width=1\linewidth]{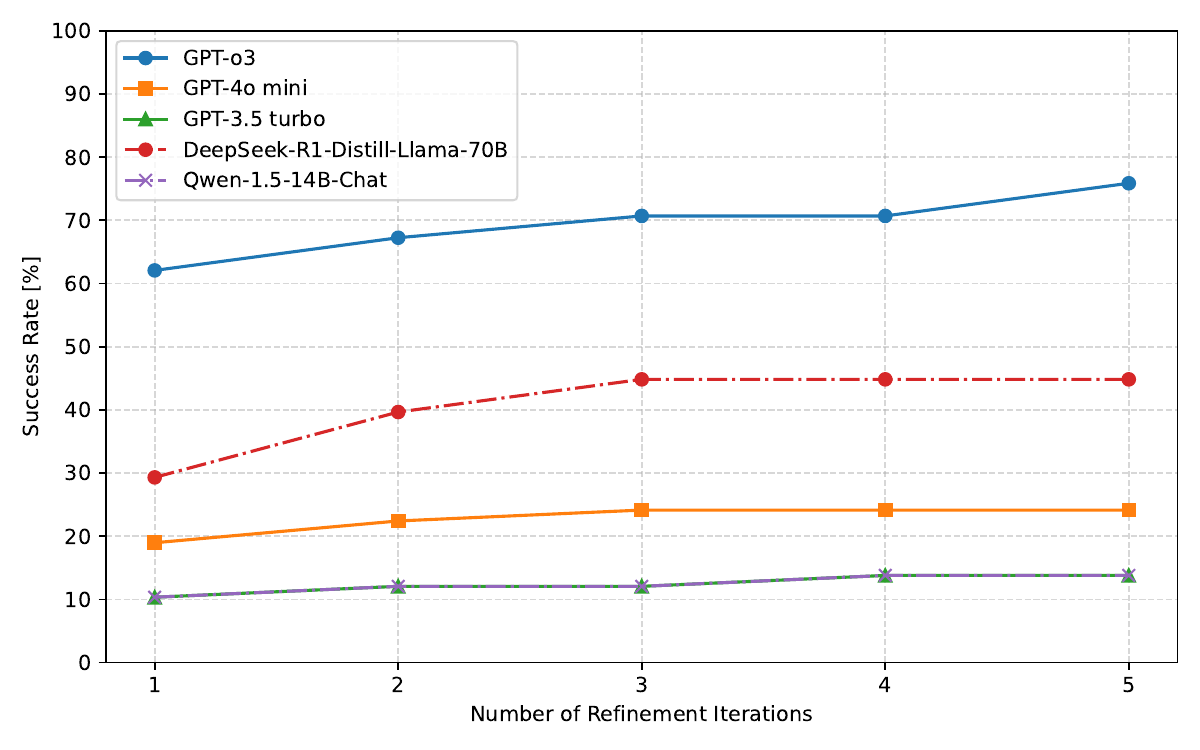}
    \caption{The changes of success rate when we change the number of refinement iterations.}
    \label{fig:iterations}
\end{figure}

As shown in Figure~\ref{fig:iterations}, the success rate improves significantly with iterative refinement across all models—GPT-3.5, GPT-4o, and o3. Notably, the most substantial gain is observed after the first refinement, particularly when increasing the iteration count from 1 to 2. Beyond the second iteration, the improvements become more incremental, indicating diminishing returns. These results suggest that leveraging feedback from the hardware-aware evaluation tool to refine the generated code is highly effective, especially in the early iterations.

\subsection*{How LLMs' performances against Human-written Submissions?}

The best-performing model, o3, achieves a success rate of 65.52\%, which significantly surpasses the averaged human performance of 39.98\%.
In contrast, GPT-4o-mini (18.97\%) and GPT-3.5-Turbo (10.34\%) perform notably worse than human coders, indicating that mid-tier LLMs still fall short in quantum programming tasks.

\begin{figure}
    \centering
    \includegraphics[width=1.0\linewidth]{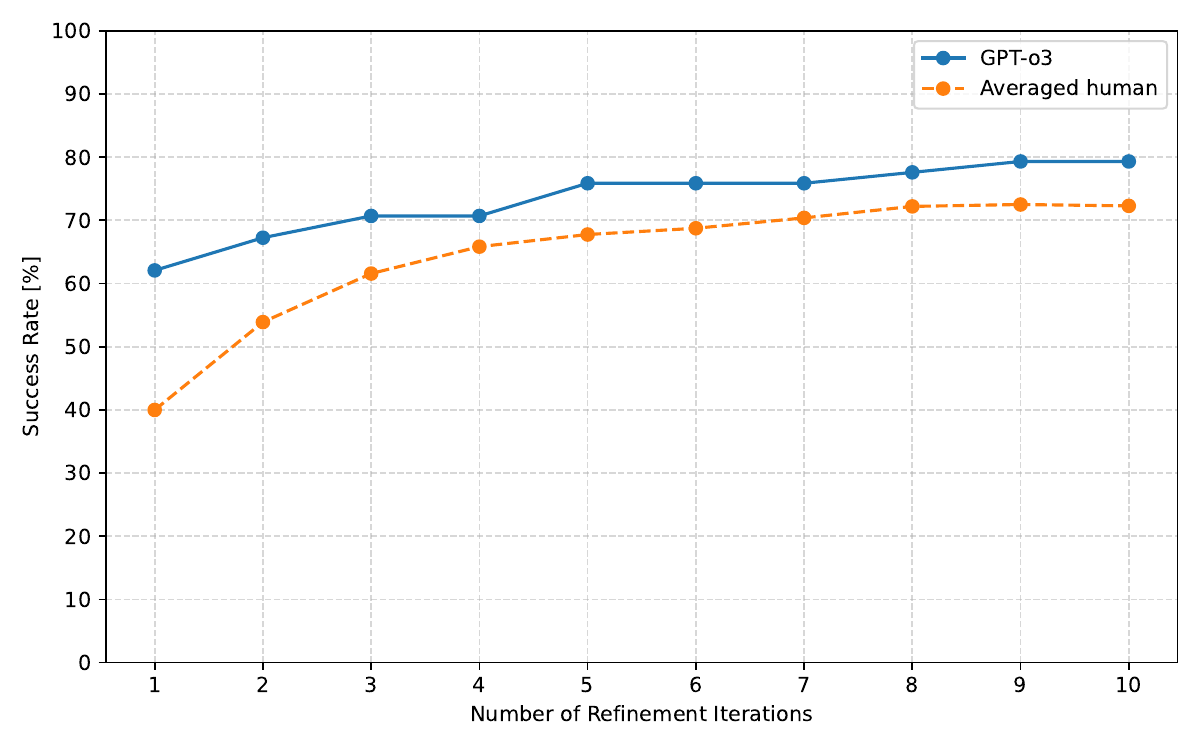}
    \caption{Changes of success rate of the best performing LLM (o3) and human submissions when we increase the number of code refinement. o3 achieves higher success rates than values obtained by averaging human success rates. }
    \label{fig:o3_vs_human}
\end{figure}

Figure~\ref{fig:o3_vs_human} illustrates the comparison between the success rates of the best-performing LLM (o3) and the averaged human performance, across different numbers of refinement iterations ranging from 1 to 15.
Note that both blue lines in Figure~\ref{fig:iterations} and Figure~\ref{fig:o3_vs_human} represent the performances of the same model.
The orange line represents the success rate of averaged human submissions.
While the human performance shows a steady upward trend as the number of refinements increases, o3 occasionally exhibits sudden gains in performance—for instance, between iteration 4 and 5.
However, it is worth noting that the success rate of o3 does not always improve monotonically; in some cases (e.g., from iteration 3 to 4), performance may stagnate or even slightly drop.
This fluctuation highlights the non-deterministic nature of LLM-based generation and suggests that, although o3 generally outperforms human coders in our datasets, its iterative refinement process is not always consistently effective.


\section{Case Study: Actual Outputs}

\begin{figure}[htb]
\centering
\begin{tabular}{|p{0.95\linewidth}|}
\hline
\textbf{Problem:} Implement the operation of preparing the state $\ket{\psi}$ from the zero state on a quantum circuit $\mathrm{qc}$ with $2$ qubits. The state $\ket{\psi}$ is defined as\\
\begin{equation}
\ket{\psi} = a_0\ket{00} + a_1\ket{10} + a_2\ket{01}, \nonumber
\end{equation}
where $a_0$, $a_1$, and $a_2$ denote arbitrary non-zero probability amplitudes (any values are permitted).
 \\
\textbf{Constraints:} Global phase is ignored in judge. \\
\textbf{Code template: (omitted)} \\


\hline
\end{tabular}
\caption{An example of the problems in QCoder (QPC001-A4)}
\label{fig: QPC001_A4}
\end{figure}

This section presents a specific example from the QCoder Benchmark to illustrate how feedback can effectively improve code generation.
The example problem is shown in Figure~\ref{fig: QPC001_A4}.

\begin{figure}[ht]
\centering
\begin{tabular}{|p{0.95\linewidth}|}
\hline
\begin{verbatim}
from qiskit import QuantumCircuit

def solve() -> QuantumCircuit:
    qc = QuantumCircuit(2) 
    qc.h(0)
    qc.ch(0, 1)
    qc.cx(1, 0)
    return qc
\end{verbatim} \\
\hline
\end{tabular}
\caption{An example of the answer for QPC001-A4}
\label{fig: QPC001_A4_answer}
\end{figure}

We show the reference program for this problem in Figure \ref{fig: QPC001_A4_answer}.
In this problem, it is not necessary to consider the amplitudes of the ground states. Instead, it suffices to view the ground state as a superposition formed by dividing the initial state $\ket{00}$ into three components. This transformation can be implemented through the following steps.

First, applying a Hadamard gate to the first qubit yields the transformation
\begin{equation}
\ket{00} \xrightarrow{H(0)} \frac{1}{\sqrt{2}} (\ket{00} + \ket{10}).
\end{equation}

Next, a controlled-Hadamard gate is applied, targeting the second qubit and controlled by the first. This results in
\begin{equation}
\frac{1}{\sqrt{2}} (\ket{00} + \ket{10}) \xrightarrow{CH(0, 1)} \frac{1}{\sqrt{2}} \ket{00} + \frac{1}{2}(\ket{10} + \ket{11}).
\end{equation}

Finally, a controlled-$X$ gate is applied, targeting the first qubit and controlled by the second, producing
\begin{equation}
\begin{split}
&\frac{1}{\sqrt{2}} \ket{00} + \frac{1}{2}(\ket{10} + \ket{11})\\ 
&\xrightarrow{CX(1, 0)} \frac{1}{\sqrt{2}} \ket{00} + \frac{1}{2}(\ket{10} + \ket{01}).
\end{split}
\end{equation}

\begin{figure}[ht]
\centering
\begin{tabular}{|p{0.95\linewidth}|}
\hline
\begin{verbatim}
def solve() -> QuantumCircuit:
    qc = QuantumCircuit(2)
    a0 = 1
    a1 = 1
    a2 = 1
    norm = np.sqrt(np.abs(a0)**2 
        + np.abs(a1)**2 + np.abs(a2)**2)
    state = np.array([a0, a2, a1, 0], 
        dtype=complex) / norm
    qc.initialize(state, [0, 1])
    return qc
\end{verbatim} \\
\hline
\end{tabular}
\caption{DeepSeek-R1's answer at the first iteration without refinement.}
\vspace{-3ex}
\label{fig: deepseek_answer_1}
\end{figure}

\begin{figure}[ht]
\centering
\begin{tabular}{|p{0.95\linewidth}|}
\hline
\begin{verbatim}
def solve() -> QuantumCircuit:
    qc = QuantumCircuit(2)
    a0 = 1
    a1 = 1
    a2 = 1
    norm = math.sqrt(a0**2+a1**2+a2**2)
    a0_norm = a0 / norm
    a1_norm = a1 / norm
    a2_norm = a2 / norm
    denominator = math.sqrt(a0_norm**2 
        + a2_norm**2)
    theta = 2 * math.atan(a1_norm 
        / denominator)
    qc.ry(theta, 0)
    phi = 2 * math.atan(a2_norm/a0_norm)
    c_ry = RYGate(phi).control(1)
    qc.append(c_ry, [0, 1])
    return qc
\end{verbatim} \\
\hline
\end{tabular}
\caption{DeepSeek-R1's answer at second iteration.}
\vspace{-3ex}
\label{fig: deepseek_answer_2}
\end{figure}

Through this sequence of operations, a superposition of the three computational basis states $\ket{00}$, $\ket{10}$, and $\ket{01}$ is successfully prepared. The available code is shown in Fig. \ref{fig: QPC001_A4_answer}.

Fig. \ref{fig: deepseek_answer_1} shows the code generated at the first iteration.
In this code, the \texttt{.initialize()} method is used, however, since this method is not included in the allowed set of quantum gates.
Therefore, the gate constraint violation error is raised. This violation report is then included in the prompt for refinement as shown in Section \ref{sec:methods},
The second code generated by the model is shown in Fig. \ref{fig: deepseek_answer_2}.
This code passes the runtime error check, however, the evaluator produces the wrong output error although the runtime or other constraints' checks are passed.
The generated code at third iteration shown in Fig. \ref{fig: deepseek_answer_3} can successfully generate the desired state. Note that the model employs a different approach from the example solution shown in Fig. \ref{fig: QPC001_A4_answer}.
Through iterative refinement, the model gradually modifies the code to produce one that satisfies the input constraints.

Finally, we also observed that many failure cases involved Python's runtime errors that are relatively easy for humans to fix, such as missing import statements (e.g., import math).
They still struggle with basic coding hygiene that would typically be trivial for humans.

\begin{figure}[ht]
\centering
\begin{tabular}{|p{0.95\linewidth}|}
\hline
\begin{verbatim}
import math
from qiskit import QuantumCircuit
from qiskit.circuit.library.
    standard_gates import RYGate

def solve() -> QuantumCircuit:
    qc = QuantumCircuit(2)
    a0 = 1
    a1 = 1
    a2 = 1
    norm = math.sqrt(a0**2+a1**2+a2**2)
    a0_norm = a0 / norm
    a1_norm = a1 / norm
    a2_norm = a2 / norm

    theta = 2 * math.asin(a1_norm)
    phi = 2*math.atan(a0_norm/a2_norm)

    qc.ry(theta, 0)
    qc.x(0)
    qc.cry(phi, 0, 1)
    qc.x(0)
    return qc
\end{verbatim} \\
\hline
\end{tabular}
\caption{DeepSeek-R1's answer at third iteration. (Success)}
\vspace{-3ex}
\label{fig: deepseek_answer_3}
\end{figure}

\section{Conclusions}

We introduced QCoder Benchmark, which consists of a dataset and evaluation tool for quantum code generation.
This benchmark is designed to investigate the capabilities of LLMs under domain-specific hardware-aware constraints.
By integrating a quantum simulator that returns hardware-aware feedback, we implemented a feedback-driven iterative code generator.
Our experiments revealed that even advanced models like GPT-4o struggle with quantum programming tasks, while reasoning-oriented models like o3 show superior performance and can even outperform human-written code submissions for programming contests.
These findings suggest the importance of refinement of codes by domain-specific feedback.
While our benchmark and experiments focus on quantum programming, the proposed feedback-driven generation framework—where domain-specific constraint violations are systematically detected and incorporated into iterative code refinement—could generalize to other domains that impose strict execution constraints.
Potential applications include robotics and embedded system programming, where functional correctness alone is insufficient and compliance with real-world constraints (e.g., timing, resource usage, device compatibility) must be enforced.
We leave the exploration of such domains to future work.
\section*{Ethical Considerations}

This work evaluates LLMs on quantum code generation using a custom benchmark dataset and simulator-based feedback. The dataset includes human-written quantum programming codes, which were collected with permission from the QCoder platform. All collected data are free of personally identifiable information and originate from programming contest submissions.

While the dataset is not yet publicly released, we plan to make it available for academic research use under a license that prohibits commercial use. The dataset and evaluation API will be distributed to promote transparency, reproducibility, and responsible research in quantum programming. Final license terms will be announced at the time of release.

\vspace{-1ex}
\section*{Acknowledgements}
\vspace{-1ex}

This work was supported in part by the Council for Science, Technology and Innovation (CSTI) through the Cross-ministerial Strategic Innovation Promotion Program (SIP), “Promoting the application of advanced quantum technology platforms to social issues” (Funding agency: QST), and by the AIST policy-based budget project “R\&D on Generative AI Foundation Models for the Physical Domain.”

\bibliography{custom}

\begin{thebibliography}{19}
\providecommand{\natexlab}[1]{#1}

\bibitem[{Austin et~al.(2021)Austin, Odena, Nye, Bosma, Michalewski, Dohan, Jiang, Cai, Terry, Le et~al.}]{austin2021program}
Jacob Austin, Augustus Odena, Maxwell Nye, Maarten Bosma, Henryk Michalewski, David Dohan, Ellen Jiang, Carrie Cai, Michael Terry, Quoc Le, and 1 others. 2021.
\newblock Program synthesis with large language models.
\newblock \emph{arXiv preprint arXiv:2108.07732}.

\bibitem[{Bai et~al.(2023)Bai, Bai, Chu, Cui, Dang, Deng, Fan, Ge, Han, Huang, Hui, Ji, Li, Lin, Lin, Liu, Liu, Lu, Lu, Ma, Men, Ren, Ren, Tan, Tan, Tu, Wang, Wang, Wang, Wu, Xu, Xu, Yang, Yang, Yang, Yang, Yao, Yu, Yuan, Yuan, Zhang, Zhang, Zhang, Zhang, Zhou, Zhou, Zhou, and Zhu}]{qwen}
Jinze Bai, Shuai Bai, Yunfei Chu, Zeyu Cui, Kai Dang, Xiaodong Deng, Yang Fan, Wenbin Ge, Yu~Han, Fei Huang, Binyuan Hui, Luo Ji, Mei Li, Junyang Lin, Runji Lin, Dayiheng Liu, Gao Liu, Chengqiang Lu, Keming Lu, and 29 others. 2023.
\newblock Qwen technical report.
\newblock \emph{arXiv preprint arXiv:2309.16609}.

\bibitem[{Bi et~al.(2024)Bi, Wan, Wang, Zhang, Guan, Lu, Zhang, Sui, Jin, and Shi}]{bi-etal-2024-iterative}
Zhangqian Bi, Yao Wan, Zheng Wang, Hongyu Zhang, Batu Guan, Fangxin Lu, Zili Zhang, Yulei Sui, Hai Jin, and Xuanhua Shi. 2024.
\newblock \href {https://doi.org/10.18653/v1/2024.findings-acl.138} {Iterative refinement of project-level code context for precise code generation with compiler feedback}.
\newblock In \emph{Findings of the Association for Computational Linguistics: ACL 2024}, pages 2336--2353, Bangkok, Thailand. Association for Computational Linguistics.

\bibitem[{Chen et~al.(2021)Chen, Tworek, Jun, Yuan, de~Oliveira~Pinto, Kaplan, Edwards, Burda, Joseph, Brockman, Ray, Puri, Krueger, Petrov, Khlaaf, Sastry, Mishkin, Chan, Gray, Ryder, Pavlov, Power, Kaiser, Bavarian, Winter, Tillet, Such, Cummings, Plappert, Chantzis, Barnes, Herbert-Voss, Guss, Nichol, Paino, Tezak, Tang, Babuschkin, Balaji, Jain, Saunders, Hesse, Carr, Leike, Achiam, Misra, Morikawa, Radford, Knight, Brundage, Murati, Mayer, Welinder, McGrew, Amodei, McCandlish, Sutskever, and Zaremba}]{chen2021codex}
Mark Chen, Jerry Tworek, Heewoo Jun, Qiming Yuan, Henrique~Ponde de~Oliveira~Pinto, Jared Kaplan, Harri Edwards, Yuri Burda, Nicholas Joseph, Greg Brockman, Alex Ray, Raul Puri, Gretchen Krueger, Michael Petrov, Heidy Khlaaf, Girish Sastry, Pamela Mishkin, Brooke Chan, Scott Gray, and 39 others. 2021.
\newblock \href {https://arxiv.org/abs/2107.03374} {Evaluating large language models trained on code}.

\bibitem[{DeepSeek-AI(2025)}]{deepseekai2025}
DeepSeek-AI. 2025.
\newblock \href {https://arxiv.org/abs/2501.12948} {Deepseek-r1: Incentivizing reasoning capability in llms via reinforcement learning}.
\newblock \emph{Preprint}, arXiv:2501.12948.

\bibitem[{Ding et~al.(2024)Ding, Min, Kaiser, and Ray}]{refine_code}
Yangruibo Ding, Marcus~J. Min, Gail Kaiser, and Baishakhi Ray. 2024.
\newblock \href {https://doi.org/10.1145/3649825} {Cycle: Learning to self-refine the code generation}.
\newblock \emph{Proc. ACM Program. Lang.}, 8(OOPSLA1).

\bibitem[{Hendrycks et~al.(2021)Hendrycks, Basart, Kadavath, Mazeika, Arora, Guo, Burns, Puranik, He, Song, and Steinhardt}]{hendrycksapps2021}
Dan Hendrycks, Steven Basart, Saurav Kadavath, Mantas Mazeika, Akul Arora, Ethan Guo, Collin Burns, Samir Puranik, Horace He, Dawn Song, and Jacob Steinhardt. 2021.
\newblock Measuring coding challenge competence with apps.
\newblock \emph{NeurIPS}.

\bibitem[{Javadi-Abhari et~al.(2024)Javadi-Abhari, Treinish, Krsulich, Wood, Lishman, Gacon, Martiel, Nation, Bishop, Cross, Johnson, and Gambetta}]{qiskit2024}
Ali Javadi-Abhari, Matthew Treinish, Kevin Krsulich, Christopher~J. Wood, Jake Lishman, Julien Gacon, Simon Martiel, Paul~D. Nation, Lev~S. Bishop, Andrew~W. Cross, Blake~R. Johnson, and Jay~M. Gambetta. 2024.
\newblock \href {https://doi.org/10.48550/arXiv.2405.08810} {Quantum computing with {Q}iskit}.
\newblock \emph{Preprint}, arXiv:2405.08810.

\bibitem[{Lai et~al.(2022)Lai, Li, Wang, Zhang, Zhong, Zettlemoyer, Yih, Fried, Wang, and Yu}]{Lai2022DS1000}
Yuhang Lai, Chengxi Li, Yiming Wang, Tianyi Zhang, Ruiqi Zhong, Luke Zettlemoyer, Wen-Tau Yih, Daniel Fried, Sida Wang, and Tao Yu. 2022.
\newblock Ds-1000: A natural and reliable benchmark for data science code generation.
\newblock \emph{ArXiv}, abs/2211.11501.

\bibitem[{Liu et~al.(2023)Liu, Zhu, Xiao, FU, Han, Wei, and Ye}]{liu2023rltf}
Jiate Liu, Yiqin Zhu, Kaiwen Xiao, QIANG FU, Xiao Han, Yang Wei, and Deheng Ye. 2023.
\newblock \href {https://openreview.net/forum?id=hjYmsV6nXZ} {{RLTF}: Reinforcement learning from unit test feedback}.
\newblock \emph{Transactions on Machine Learning Research}.

\bibitem[{Luo et~al.(2024)Luo, Wu, Liu, and Antwi-Afari}]{LUO2024100488}
Hanbin Luo, Jianxin Wu, Jiajing Liu, and Maxwell~Fordjour Antwi-Afari. 2024.
\newblock \href {https://doi.org/10.1016/j.dibe.2024.100488} {Large language model-based code generation for the control of construction assembly robots: A hierarchical generation approach}.
\newblock \emph{Developments in the Built Environment}, 19:100488.

\bibitem[{Madaan et~al.(2023)Madaan, Tandon, Gupta, Hallinan, Gao, Wiegreffe, Alon, Dziri, Prabhumoye, Yang, Welleck, Majumder, Gupta, Yazdanbakhsh, and Clark}]{madaan2023selfrefine}
Aman Madaan, Niket Tandon, Prakhar Gupta, Skyler Hallinan, Luyu Gao, Sarah Wiegreffe, Uri Alon, Nouha Dziri, Shrimai Prabhumoye, Yiming Yang, Sean Welleck, Bodhisattwa~Prasad Majumder, Shashank Gupta, Amir Yazdanbakhsh, and Peter Clark. 2023.
\newblock \href {https://arxiv.org/abs/2303.17651} {Self-refine: Iterative refinement with self-feedback}.
\newblock \emph{Preprint}, arXiv:2303.17651.

\bibitem[{OpenAI et~al.(2024)OpenAI, Achiam, Adler, Agarwal, Ahmad, Akkaya, Aleman, Almeida, Altenschmidt, Altman, Anadkat, Avila, Babuschkin, Balaji, Balcom, Baltescu, Bao, Bavarian, Belgum, Bello, Berdine, Bernadett-Shapiro, Berner, Bogdonoff, Boiko, Boyd, Brakman, Brockman, Brooks, Brundage, Button, Cai, Campbell, Cann, Carey, Carlson, Carmichael, Chan, Chang, Chantzis, Chen, Chen, Chen, Chen, Chen, Chess, Cho, Chu, Chung, Cummings, Currier, Dai, Decareaux, Degry, Deutsch, Deville, Dhar, Dohan, Dowling, Dunning, Ecoffet, Eleti, Eloundou, Farhi, Fedus, Felix, Fishman, Forte, Fulford, Gao, Georges, Gibson, Goel, Gogineni, Goh, Gontijo-Lopes, Gordon, Grafstein, Gray, Greene, Gross, Gu, Guo, Hallacy, Han, Harris, He, Heaton, Heidecke, Hesse, Hickey, Hickey, Hoeschele, Houghton, Hsu, Hu, Hu, Huizinga, Jain, Jain, Jang, Jiang, Jiang, Jin, Jin, Jomoto, Jonn, Jun, Kaftan, Łukasz Kaiser, Kamali, Kanitscheider, Keskar, Khan, Kilpatrick, Kim, Kim, Kim, Kirchner, Kiros, Knight, Kokotajlo, Łukasz Kondraciuk,
  Kondrich, Konstantinidis, Kosic, Krueger, Kuo, Lampe, Lan, Lee, Leike, Leung, Levy, Li, Lim, Lin, Lin, Litwin, Lopez, Lowe, Lue, Makanju, Malfacini, Manning, Markov, Markovski, Martin, Mayer, Mayne, McGrew, McKinney, McLeavey, McMillan, McNeil, Medina, Mehta, Menick, Metz, Mishchenko, Mishkin, Monaco, Morikawa, Mossing, Mu, Murati, Murk, Mély, Nair, Nakano, Nayak, Neelakantan, Ngo, Noh, Ouyang, O'Keefe, Pachocki, Paino, Palermo, Pantuliano, Parascandolo, Parish, Parparita, Passos, Pavlov, Peng, Perelman, de~Avila Belbute~Peres, Petrov, de~Oliveira~Pinto, Michael, Pokorny, Pokrass, Pong, Powell, Power, Power, Proehl, Puri, Radford, Rae, Ramesh, Raymond, Real, Rimbach, Ross, Rotsted, Roussez, Ryder, Saltarelli, Sanders, Santurkar, Sastry, Schmidt, Schnurr, Schulman, Selsam, Sheppard, Sherbakov, Shieh, Shoker, Shyam, Sidor, Sigler, Simens, Sitkin, Slama, Sohl, Sokolowsky, Song, Staudacher, Such, Summers, Sutskever, Tang, Tezak, Thompson, Tillet, Tootoonchian, Tseng, Tuggle, Turley, Tworek, Uribe, Vallone,
  Vijayvergiya, Voss, Wainwright, Wang, Wang, Wang, Ward, Wei, Weinmann, Welihinda, Welinder, Weng, Weng, Wiethoff, Willner, Winter, Wolrich, Wong, Workman, Wu, Wu, Wu, Xiao, Xu, Yoo, Yu, Yuan, Zaremba, Zellers, Zhang, Zhang, Zhao, Zheng, Zhuang, Zhuk, and Zoph}]{openai2024gpt4technicalreport}
OpenAI, Josh Achiam, Steven Adler, Sandhini Agarwal, Lama Ahmad, Ilge Akkaya, Florencia~Leoni Aleman, Diogo Almeida, Janko Altenschmidt, Sam Altman, Shyamal Anadkat, Red Avila, Igor Babuschkin, Suchir Balaji, Valerie Balcom, Paul Baltescu, Haiming Bao, Mohammad Bavarian, Jeff Belgum, and 262 others. 2024.
\newblock \href {https://arxiv.org/abs/2303.08774} {Gpt-4 technical report}.
\newblock \emph{Preprint}, arXiv:2303.08774.

\bibitem[{Papineni et~al.(2002)Papineni, Roukos, Ward, and Zhu}]{papineni-etal-2002-bleu}
Kishore Papineni, Salim Roukos, Todd Ward, and Wei-Jing Zhu. 2002.
\newblock \href {https://doi.org/10.3115/1073083.1073135} {{B}leu: a method for automatic evaluation of machine translation}.
\newblock In \emph{Proceedings of the 40th Annual Meeting of the Association for Computational Linguistics}, pages 311--318, Philadelphia, Pennsylvania, USA. Association for Computational Linguistics.

\bibitem[{Tang et~al.(2024)Tang, Qian, Gao, Chen, Chen, and Gerstein}]{10.1093/bioinformatics/btae230}
Xiangru Tang, Bill Qian, Rick Gao, Jiakang Chen, Xinyun Chen, and Mark~B Gerstein. 2024.
\newblock \href {https://doi.org/10.1093/bioinformatics/btae230} {Biocoder: a benchmark for bioinformatics code generation with large language models}.
\newblock \emph{Bioinformatics}, 40(Supplement-1):i266--i276.

\bibitem[{Tony et~al.(2023)Tony, Mutas, Díaz~Ferreyra, and Scandariato}]{llmseceval2023}
Catherine Tony, Markus Mutas, Nicolas Díaz~Ferreyra, and Riccardo Scandariato. 2023.
\newblock \href {https://doi.org/10.5281/zenodo.7565965} {Llmseceval: A dataset of natural language prompts for security evaluations}.
\newblock In \emph{2023 IEEE/ACM 20th International Conference on Mining Software Repositories (MSR)}.

\bibitem[{Vishwakarma et~al.(2024)Vishwakarma, Harkins, Golecha, Bajpe, Dupuis, Buratti, Kremer, Faro, Puri, and Cruz-Benito}]{vishwakarma2024qiskithumaneval}
Sanjay Vishwakarma, Francis Harkins, Siddharth Golecha, Vishal~Sharathchandra Bajpe, Nicolas Dupuis, Luca Buratti, David Kremer, Ismael Faro, Ruchir Puri, and Juan Cruz-Benito. 2024.
\newblock \href {https://arxiv.org/abs/2406.14712} {Qiskit humaneval: An evaluation benchmark for quantum code generative models}.
\newblock \emph{Preprint}, arXiv:2406.14712.

\bibitem[{Wang et~al.(2025)Wang, Li, Song, Xu, Tang, Zhuge, Pan, Song, Li, Singh, Tran, Li, Ma, Zheng, Qian, Shao, Muennighoff, Zhang, Hui, Lin, Brennan, Peng, Ji, and Neubig}]{wang2025openhands}
Xingyao Wang, Boxuan Li, Yufan Song, Frank~F. Xu, Xiangru Tang, Mingchen Zhuge, Jiayi Pan, Yueqi Song, Bowen Li, Jaskirat Singh, Hoang~H. Tran, Fuqiang Li, Ren Ma, Mingzhang Zheng, Bill Qian, Yanjun Shao, Niklas Muennighoff, Yizhe Zhang, Binyuan Hui, and 5 others. 2025.
\newblock \href {https://openreview.net/forum?id=OJd3ayDDoF} {Openhands: An open platform for {AI} software developers as generalist agents}.
\newblock In \emph{The Thirteenth International Conference on Learning Representations}.

\bibitem[{Yu et~al.(2018)Yu, Zhang, Yang, Yasunaga, Wang, Li, Ma, Li, Yao, Roman, Zhang, and Radev}]{yu-etal-2018-spider}
Tao Yu, Rui Zhang, Kai Yang, Michihiro Yasunaga, Dongxu Wang, Zifan Li, James Ma, Irene Li, Qingning Yao, Shanelle Roman, Zilin Zhang, and Dragomir Radev. 2018.
\newblock \href {https://doi.org/10.18653/v1/D18-1425} {{S}pider: A large-scale human-labeled dataset for complex and cross-domain semantic parsing and text-to-{SQL} task}.
\newblock In \emph{Proceedings of the 2018 Conference on Empirical Methods in Natural Language Processing}, pages 3911--3921, Brussels, Belgium. Association for Computational Linguistics.

\end{thebibliography}

\clearpage

\end{CJK}
\end{document}